\def\year{2020}\relax
\documentclass[letterpaper]{article} 
\usepackage{aaai20}  
\usepackage{times}  
\usepackage{helvet} 
\usepackage{courier}  
\usepackage[hyphens]{url}  
\usepackage{graphicx} 
\usepackage{graphicx}  

\usepackage{amssymb}
\usepackage{booktabs} 
\usepackage{bm}
\usepackage[ruled,lined,linesnumbered]{algorithm2e}
\usepackage{algorithmic}
\usepackage{array}
\usepackage{multirow}
\usepackage{enumerate}
\usepackage{subfigure}
\usepackage{dsfont}
\usepackage{amsfonts}
\usepackage{amsmath}
\usepackage{adjustbox}
\usepackage{blindtext}
\graphicspath{ {figures/} }
\title{Multi-source Domain Adaptation for Visual Sentiment Classification }

\author{%
	Chuang Lin\textsuperscript{\rm 1}\thanks{Equal Contribution.}, 
	Sicheng Zhao\textsuperscript{\rm 2}\footnotemark[1], 
	Lei Meng\textsuperscript{\rm 1}\thanks{Corresponding Author.}, 
	Tat-Seng Chua\textsuperscript{\rm 1}\\
	\textsuperscript{\rm 1}NExT++, National University of Singapore\\
	\textsuperscript{\rm 2}University of California, Berkeley, USA \\
	clin@u.nus.edu,
	schzhao@gmail.com,
	lmeng@nus.edu.sg,
	dcscts@nus.edu.sg
}

\begin{document}
\maketitle

\begin{abstract}
Existing domain adaptation methods on visual sentiment classification typically are investigated under the single-source scenario, where the knowledge learned from a source domain of sufficient labeled data is transferred to the target domain of loosely labeled or unlabeled data. 
However, in practice, data from a single source domain usually have a limited volume and can hardly cover the characteristics of the target domain. 
In this paper, we propose a novel multi-source domain adaptation (MDA) method, termed Multi-source Sentiment Generative Adversarial Network (MSGAN), for visual sentiment classification. 
To handle data from multiple source domains, it learns to find a unified sentiment latent space where data from both the source and target domains share a similar distribution. 
This is achieved via cycle consistent adversarial learning in an end-to-end manner. 
Extensive experiments conducted on four benchmark datasets demonstrate that MSGAN significantly outperforms the state-of-the-art MDA approaches for visual sentiment classification.
\end{abstract}

\section{Introduction}
\label{sec:Introduction}
It is increasingly popular for people to share experiences, express opinions, and record activities by posting images on social networks like Instagram and Twitter.
It offers a great opportunity for visual sentiment analysis that infers their emotional behaviors and provide personalized services \cite{zhao2018predicting}, such as blog recommendation~\cite{borth2013large} and tourism~\cite{alaei2019sentiment}.

Recent advances in deep learning have significantly improved the state-of-the-art performance in visual sentiment classification~\cite{yang2018weakly,yang2018visual,katsurai2016image} or image emotion distribution learning~\cite{yang2017joint}.
However, considering the difficulties in the acquisition of sentiment labels due to the high subjectivity in the human perception process, training a model on a labeled source domain that can well generalize to another new domain is necessary.
Because of the presence of \emph{domain shift} or \emph{dataset bias}~\cite{torralba2011unbiased}, even a slight departure from a network's training domain can lead to incorrect predictions and significantly reduce its performance.
\begin{figure}[!t]
\begin{center}
\centering \includegraphics[width=0.95\linewidth]{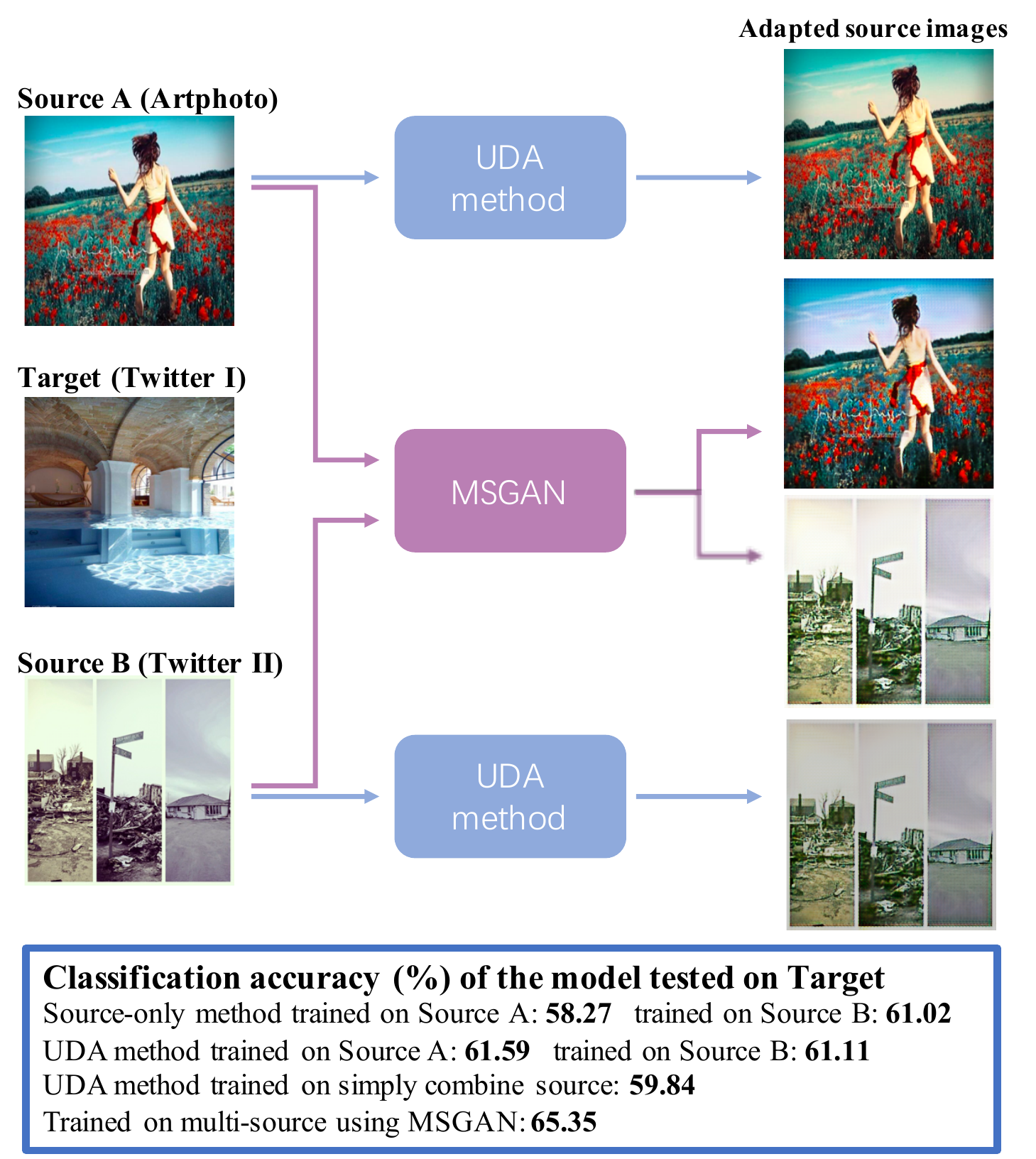}
\caption{An example of \emph{domain shift} in  sentiment classification using training data from multiple sources.}
\label{fig:DomainShift}
\end{center}
\end{figure}
Therefore, establishing knowledge transfer from a labeled source domain to an unlabeled target domain for visual sentiment analysis has attracted significant attention~\cite{zhao2018emotiongan,zhao2019cycleemotiongan}.
Recently, most existing deep unsupervised domain adaptation (UDA) methods assume that there is only one single source domain and the labeled source data are implicitly sampled from the same underlying distribution.
In practice, there are obvious biases in different domains, even for the images of the same sentiment. 
For example, as shown in Figure~\ref{fig:DomainShift}, the styles of artistic photos, abstract paintings, and natural images are quite different. 
Simply combining different sources into one source and directly employing a state-of-the-art single-source UDA model~\protect\cite{zhu2017unpaired} lead to a drop in classification accuracy from 61.59\% (if trained on the single-best source) to 59.84\% (if trained on simply combined source). 
This is because images from different source domains may interfere with each other during the learning process~\cite{riemer2018learning}. 
Comparing to source-only method~\cite{peng2015mixed}, UDA model can alleviate domain shift between single source domain and the target domain to some extent.
However, it still exists between multiple source domains and target domain, leading to multi-source domain adaptation (MDA) for visual sentiment classification. Despite of the rapid progress in UDA, no study has been investigated on MDA for visual sentiment classification.
This task is much more challenging due to the following reasons.
First, the multiple sources are different not only from the target but also from each other.
Existing MDA methods only align each source and target pair.
Although different sources are matched towards the target, there may exist significant mis-alignment across different sources.
Second, these MDA methods focus on matching the visual features but ignore the semantic labels, which hardly characterize the consistency of image sentiment.
Third, current methods typically require multiple classifiers or transfer structures for different source domains.
This leads to a high model complexity and low learning stability when learning from small-scale sentiment data.  

In this paper, we study the multi-source unsupervised domain adaptation problem for visual sentiment classification.
Specifically, we propose a novel adversarial framework, termed Multi-source Sentiment Generative Adversarial Network (MSGAN), which is composed of three pipelines, \emph{i.e.} image reconstruction, image translation, and cycle-reconstruction.
The image reconstruction and cycle-reconstruction pipelines learn a unified sentiment space where data from the source and target domains share similar distributions.
Subsequently, the image translation pipeline restricted by emotional semantic consistency learns to adapt the source domain images to appear as if they were drawn from the target domain, while preserving the annotation information. 
Notably, thanks to the unified sentiment latent space, MSGAN requires a single classification network to handle data from different source domains. 

In summary, the contributions of this paper are threefold:

1. We propose to adapt visual sentiment from multiple source domains to a target domain in an end-to-end manner. 
To the best of our knowledge, this is the first multi-source domain adaptation work on visual sentiment classification.	

2. We develop a novel adversarial framework, termed MSGAN, for visual sentiment classification. 
By the joint learning of image reconstruction, image translation, and cycle-reconstruction pipelines, images from multiple sources and one target can be mapped to a unified sentiment latent space.
Meanwhile, the semantic consistency loss in image translation pipeline preserves the semantic information of images.

3. We conduct extensive experiments on the ArtPhoto~\cite{machajdik2010affective}, FI~\cite{you2016building}, Twitter \uppercase\expandafter{\romannumeral1}~\cite{you2015robust}, and Twitter \uppercase\expandafter{\romannumeral2}~\cite{you2016building} datasets, and the results demonstrate the superiority of the proposed MSGAN model compared with the state-of-the-art MDA approaches.

\section{Related Work}

\noindent\textbf{Visual Sentiment Classification:}
Recently, with the great success of convolutional neural network (CNN) on many computer vision tasks, CNN has also been employed in sentiment classification.
\citeauthor{you2015robust}~\shortcite{you2015robust} proposed a progressive CNN architecture to make use of noisily labeled data for binary sentiment classification. 
\citeauthor{yang2018retrieving}~\shortcite{yang2018retrieving} employed deep metric learning to optimize both retrieval and classification tasks by jointly optimizing cross-entropy loss and a novel sentiment constraint. Different from improving global image representations, several methods~\cite{you2017visual,yang2018weakly} consider the local information for sentiment classification.
All the above methods employ a supervised manner to learn the mapping between image content and sentiments. 
In this paper, we study how to adapt the models from multiple labeled source domain to an unlabeled target domain for visual sentiment classification.

\noindent\textbf{Single-source Domain Adaptation:}
Since data from the source and target domains have intrinsically different distributions, the key problem in single-source UDA is how to reduce the domain shift.
Discrepancy-based methods explicitly measure the discrepancy between the source and target domains on corresponding activation layers of the two network streams~\cite{sun2017correlation,zhuo2017deep};
Adversarial generative models combine the domain discriminative model with a generative component generally based on GANs~\cite{goodfellow2014generative}. 
The Coupled Generative Adversarial Networks (CoGAN)~\cite{liu2016coupled} can learn a joint distribution of multi-domain images with a tuple of GANs;
Reconstruction based methods incorporate a reconstruction loss to minimize the difference between the input and the reconstructed input~\cite{ghifary2015domain,ghifary2016deep}. 
\citeauthor{zhao2019cycleemotiongan}~\shortcite{zhao2019cycleemotiongan} proposed CycleEmotionGAN for image emotion classification by adapting source domain images to have similar distributions to the target ones by enforcing emotional semantic consistency.
However, none of them can handle data from multiple source domains, which is the target of this paper.

\noindent\textbf{Multi-source Domain Adaptation:}
Compared with single source UDA, multi-source domain adaptation (MDA) assumes that training data from multiple sources are available~\cite{zhao2019multi}.
Early efforts on this task used shallow models~\cite{sun2013bayesian}.
MDA also develops with theoretical supports.
\citeauthor{blitzer2008learning}~\shortcite{blitzer2008learning} provided the first learning bound for MDA. 
\citeauthor{mansour2009domain}~\shortcite{mansour2009domain} claimed that an ideal target hypothesis can be represented by a distribution of a weighted combination of source hypotheses. 
In the more applied works, Deep Cocktail Network (DCTN)~\cite{xu2018deep} proposed a k-way domain discriminator and category classifier for digit classification and real-world object recognition.
\citeauthor{zhao2018adversarial}~\shortcite{zhao2018adversarial} proposed new generalization bounds and algorithms under both classification and regression settings for MDA.
\citeauthor{peng2018moment}~\shortcite{peng2018moment} directly matched all the distributions based on moments and provided a concrete proof of why matching the moments of multiple distributions works for MDA.
Different from these methods, we learn a unified sentiment latent space which jointly aligns data from all source and target domains.

\section{Problem Setup}
Suppose we have $M$ source domains $S_1$ , $S_2$ , $\cdots$, $S_M$ and one target domain $T$.
In the unsupervised multi-source domain adaptation (MDA) scenario, $S_1$ , $S_2$ , $\cdots$, $S_M$ are labeled and $T$ is fully unlabeled.
For the $i$th source domain $S_i$, the observed images and corresponding sentiment labels drawn from the source distribution $p_i(\textbf{x}, \textbf{y})$ are $X_i = \{\textbf{x}_i^j\}^{N_i}_{j=1} $ and $Y_i = \{\textbf{y}_i^j\}^{N_i}_{j=1} $,where $N_i$ is the number of images in $S_i$.
The target images drawn from the target distribution $p_T (\textbf{x}, \textbf{y})$ are $X_T = \{\textbf{x}_T^j\}^{N_T}_{j=1} $ without label observation, where $N_T$ is the number of target images.

\begin{figure}[!t]
\begin{center}
\centering \includegraphics[width=0.8\linewidth]{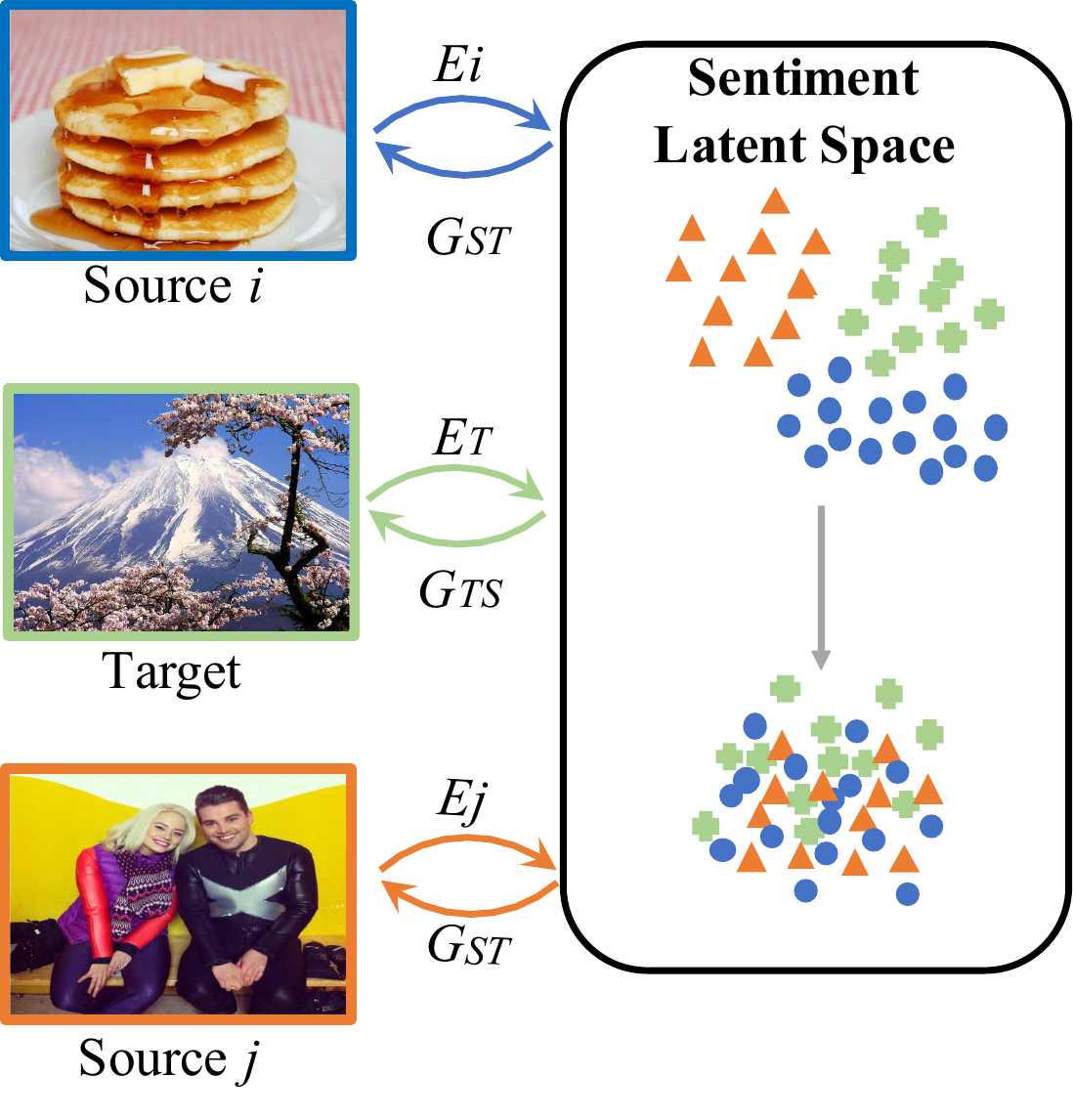}
\caption{The unified sentiment latent space. MSGAN can map source $i$, source $j$ and target domain images to a unified sentiment latent space. $E_{i}$,$E_j$ and $E_T$ are encoding functions, mapping images to latent codes. $G_{ST}$ and $G_{TS}$ are two generation functions, mapping latent codes to images.}
\label{fig:latentspace}
\end{center}
\end{figure}

The main idea of MSGAN is to learn a mapping that can align the images from both the multiple source and target domains to have similar distributions in a unified sentiment space.
As shown in Figure~\ref{fig:latentspace}, images from both the source and target domains are mapped to have similar distributions, and their information is preserved by the reconstruction loss.

To achieve this, we introduce the encoders $E_{S_1}$, $E_{S_2}$, $\cdots$, $E_{S_M}$, $E_{T}$ and generators $G_{ST}$, $G_{TS}$.
To obtain the sentiment latent space, we enforce a weight-sharing constraint on the encoders and generators. 
Specifically, we share the weights of the last block layers of $E_{S_1}$, $E_{S_2}$, $\cdots$, $E_{S_M}$ and $E_{T}$ to extract the high-level representations of the input images from the multiple sources and target. 
Similarly, we share the weights of the first block layers of $G_{ST}$ and $G_{TS}$ to decode their high-level representations for reconstructing the input images.
The unified sentiment space is learned from the cycle-consistency constrain~\cite{liu2017unsupervised}: $\textbf{x}_{i} = F^*_{T\rightarrow S}(F^*_{S\rightarrow T}(\textbf{x}_{i}))$ and $\textbf{x}_{T} = F^*_{S\rightarrow T}(F^*_{T\rightarrow S}(\textbf{x}_{T}))$, where $ F^*_{S\rightarrow T}(\textbf{x}_{i}) = G_{ST}(E_{S_i}(\textbf{x}_{i}))$, $ F^*_{T\rightarrow S}(\textbf{x}_{T}) = G_{TS}(E_{T}(\textbf{x}_{T}))$.
With the latent space established, we do not need multiple transfer structures.
\section{Multi-source Sentiment Generative Adversarial Network}
\begin{figure*}[!t]
	\begin{center}
		\centering \includegraphics[width=0.9\linewidth]{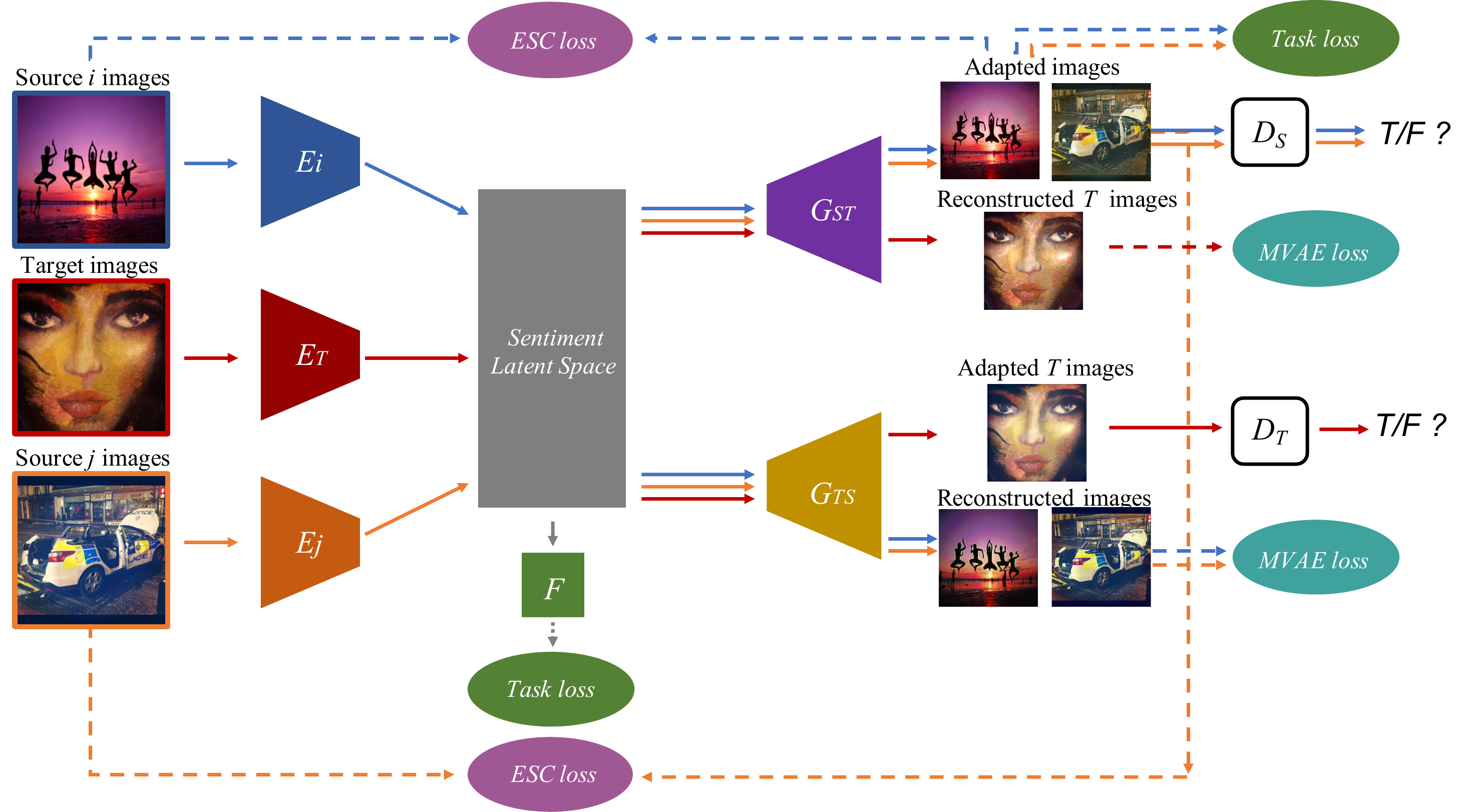}
		\caption{The framework of the proposed Multi-source Sentiment Generative Adversarial Network (MSGAN) for visual sentiment classification.The solid arrowed lines indicate the operations in the training stage. The dot dash arrowed lines correspond to different losses. For clarity the cycle-reconstruction pipeline is omitted. In the testing phase, testing data from the target domain are forwarded through the target encoder and classifier to make the final prediction.}
		\label{fig:Framework}
	\end{center}
\end{figure*}

Our framework, as illustrated in Figure~\ref{fig:Framework}, is based on variational autoencoders (VAEs) and generative adversarial networks (GANs).
It consists of three pipelines: image reconstruction, image translation, and cycle-reconstruction.
Image reconstruction pipeline includes multiple domain image encoders $E_{S_1}$, $\cdots$, $E_{S_M}$, $E_{T}$ encoding images to a unified sentiment space and two image generators $G_{ST}$, $G_{TS}$ reconstructing input images.
Image translation pipeline includes two generative adversarial networks: $GAN_{ST} = \{D_T, G_{ST}\}$ and $GAN_{TS} = \{D_S, G_{TS}\}$ learning the mapping between multiple source domains and the target domain.
The cycle reconstruction pipeline is used to learn a unified sentiment space and ensure that features of images from different domains preserved the information of their original images. 

\subsection{Image Reconstruction Pipeline}
The image reconstruction pipeline is achieved by multiple encoder$-$generator pairs $\{E_{S_i}, G_{ST}\}$ , each of which maps
an input image $\textbf{x}_i$ to a code in a latent space $Z$ via $E_{S_i}$ and then decodes a random-perturbed version of the code to reconstruct the input image via $G_{ST}$.
We assume the components in the latent space $Z$ are conditionally independent Gaussian distributions with unit variance.
The encoder outputs the latent code $\textbf{z}_i = E_{S_i}(\textbf{x}_i)$.
The reconstructed image is $\textbf{x}^\prime_{i} = G_{TS}(\textbf{z}_i, \eta)$, where $\eta$ is a noise has the same distribution of $\textbf{z}_i$.
Similarly, $\{E_{T}, G_{TS}\}$ constitutes a VAE for the target domain, where the reconstructed image is $\textbf{x}^\prime_{T} = G_{ST}(E_T(\textbf{x}_T), \eta)$.

MVAE training aims to minimize a variational upper bound as follows:
\begin{equation}
\small
\begin{aligned}
\mathcal{L}_{MVAE_i}(E_{S_i}, G_{TS}) =\lambda_0KL(q_i(\textbf{z}_i\mid\textbf{x}_i)\|p_\eta(z))-&\\
\lambda_1\mathbb{E}_{\textbf{z}_i\sim q_i(\textbf{z}_i\mid\textbf{x}_i)}[\log p_{G_{TS}}(\textbf{x}_i\mid \textbf{z}_i)],&\\
\end{aligned}
\label{equ:mvae_loss_s}
\end{equation}

\begin{equation}
\small
\begin{aligned}
\mathcal{L}_{MVAE_T}(E_T, G_{ST})
=\lambda_0KL(q_T(\textbf{z}_T\mid\textbf{x}_T)\|p_\eta(z))-&\\
\lambda_1\mathbb{E}_{\textbf{z}_T\sim q_T(\textbf{z}_T\mid\textbf{x}_T)}[\log p_{G_{ST}}(\textbf{x}_T\mid \textbf{z}_T)].&\\
\end{aligned}
\label{equ:mvae_loss_t}
\end{equation}
where the hyper-parameters $\lambda_0$ and $\lambda_1$ control the weights of the objective terms and the $KL$ divergence terms penalize the deviation of the distribution of the latent code from the prior distribution.
$p_{G_{TS}}$ and $p_{G_{ST}}$ are modeled using Laplacian distributions, respectively. 

\subsection{Image Translation Pipeline}
As aforementioned, the unified sentiment space allows to use only one generator $G_{ST}$ to adapt multi-source images indistinguishable from the target domain.
$G_{ST}$ can generate two types of images: 
(1) images from the reconstruction pipeline $\textbf{x}^\prime_{T} = G_{ST}(\textbf{z}_T\sim q_T(\textbf{z}_T\mid\textbf{x}_T))$ and 
(2) images from the translation pipeline $\textbf{x}^{S_i\rightarrow T}_{i} = G_{ST}(\textbf{z}_i\sim q_i(\textbf{z}_i\mid\textbf{x}_i))$.
A similar processing is applied to $G_{TS}$.
Meanwhile, two discriminators $D_T$ and $D_S$ are used to distinguish between $\textbf{X}_T$ and $G_{ST}(E_i(\textbf{X}_{S_i}))$, 
and $\textbf{X}_{S_i}$ and $G_{TS}(E_T(\textbf{X}_T))$, respectively.

The GAN objective functions are defined by:
 \begin{equation}
 \small
 \begin{aligned}
 \mathcal{L}_{GAN_i}(E_{S_i},G_{ST},D_T)=\mathbb{E}_{\textbf{x}_T\sim P_{X_T}}\log [1-D_T(\textbf{x}_T)]+&\\
 \mathbb{E}_{\textbf{x}_i\sim P_{X_i}}\log D_T(G_{ST}(E_{S_i}(\textbf{x}_i)),&\\
 \end{aligned}
 \label{equ:gan_loss_s}
 \end{equation}
 
 \begin{equation}
 \small
 \begin{aligned}
 \mathcal{L}_{GAN_T}(E_{T},G_{TS},D_S)=\mathbb{E}_{\textbf{x}_i\sim P_{X_i}}\log [1-D_S(\textbf{x}_i)]+&\\
 \mathbb{E}_{\textbf{x}_T\sim P_{X_T}}\log D_S(G_{TS}(E_T(\textbf{x}_T))).&\\
 \end{aligned}
 \label{equ:gan_loss_t}
 \end{equation}
 
 To preserve the semantics of the adapted images, generated by $G_{ST}$, from source to the target domain, an emotional semantic consistency loss is used, defined by:
 \begin{equation}
 \small
 \begin{aligned}
 &\mathcal{L}_{ESC}(E_{S_i},E_T,G_{ST})=\\
 &\mathbb{E}_{\textbf{x}_{S_i}\sim P_{S_i}}KL(F(E_i(\textbf{x}_{S_i}))\|F(E_T(\textbf{x}^{S_i\rightarrow T}_{i}))),
 \label{equ:esc_loss}
 \end{aligned}
 \end{equation}
 where $\textbf{x}^{S_i\rightarrow T}_{i} = G_{ST}(E_i(\textbf{x}_{i}))$ and $KL(\cdot\|\cdot)$ is the KL divergence between two distributions.
 \subsection{Cycle Reconstruction Pipeline}
 The cycle reconstruction pipeline is used to learn a unified sentiment latent space and ensure that features of images from different domains preserve the information of their original images. 
 According to~\cite{liu2017unsupervised}, a VAE-like objective function is used to model the cycle-consistency constraint, defined by:
\begin{equation}
\small
\begin{aligned}
\mathcal{L}_{cyc_i}(E_{S_i},G_{ST},E_T,G_{TS}) =\lambda_2KL(q_T(\textbf{z}_T\mid\textbf{x}^{S_i\rightarrow T}_{i} )\|\\
p_\eta(z))
-\lambda_3\mathbb{E}_{\textbf{z}_T\sim q_T(\textbf{z}_T\mid\textbf{x}^{S_i\rightarrow T}_{i} )}[\log p_{G_{TS}}(\textbf{x}_i\mid \textbf{z}_T)],
\end{aligned}
\label{equ:cyc_loss_s}
\end{equation}

 \begin{equation}
 \small
 \begin{aligned}
 \mathcal{L}_{cyc_T}(E_T,G_{TS},E_{S_i},G_{ST}) =\lambda_2KL(q_i(\textbf{z}_i\mid\textbf{x}^{T\rightarrow S_{i}}_{T} )\|\\
 p_\eta(z))
 -\lambda_3\mathbb{E}_{\textbf{z}_i\sim q_i(\textbf{z}_i\mid\textbf{x}^{T\rightarrow S_i}_{T} )}[\log p_{G_{ST}}(\textbf{x}_T\mid \textbf{z}_i)].
 \end{aligned}
 \label{equ:cyc_loss_t}
 \end{equation}
The hyper-parameters $\lambda_2$ and $\lambda_3$ control the weights of the two different objective terms.

Therefore, the augmented MVAE-GAN loss is:
\begin{equation}
\small
\begin{aligned}
&\mathcal{L}_{MVAE-GAN}(E_{S_1},\cdots,E_{S_M},E_T,G_{TS},G_{TS},D_T,D_S)\\
&=\sum_{i=1}^M\big[
\mathcal{L}_{MVAE_i}(E_{S_i}, G_{TS}) +
\mathcal{L}_{MVAE_T}(E_T, G_{ST})+\\
&\mathcal{L}_{GAN_i}(E_{S_i},G_{ST},D_T)+ 
\mathcal{L}_{cyc_i}(E_{S_i},G_{ST},E_T,G_{TS})+\\
&\mathcal{L}_{GAN_T}(E_{T},G_{TS},D_S)+
\mathcal{L}_{cyc_T}(E_T,G_{TS},E_{S_i},G_{ST})\big].\\
\end{aligned}
\label{equ:MVAE-GAN_loss}
\end{equation}

\subsection{Sentiment Classification with Adapted Images}
\label{ssec:ClassificationLoss}

After the joint learning image reconstruction, image translation, and cycle-reconstruction pipelines,
both the source images and adapted images of different domains $S_i (i=1,2,\cdots,M)$ can be mapped to a same latent representation in a unified sentiment latent space . 
Meanwhile, the semantic consistency loss in the image translation pipeline ensures the semantic information, \emph{i.e.} the corresponding sentiment labels, is preserved before and after image translation. 

Generally, multiple classification models are needed to correspond to different domains for the final classification task.
To the contrary, thanks to the unified latent space, the proposed MSGAN is augmented with a single classifier optimized by minimizing the following cross-entropy loss:

\begin{equation}
\small
\begin{aligned}
&\mathcal{L}_{task}(E_{S_i},E_T,F)=\\
&\lambda_4\mathbb{E}_{(\textbf{x}_i,y_i)\sim P_{X_i}}\sum_{l=1}^{L}\mathds{1}_{[l=y_i]}\log(\sigma(F^{(l)}(E_{S_i}(\textbf{x}_i))))\\
&+\lambda_5\mathbb{E}_{(\textbf{x}_i,y_i)\sim P_{X_i}}\sum_{l=1}^{L}\mathds{1}_{[l=y_i]}\log(\sigma(F^{(l)}(E_T(\textbf{x}_i^{S_i\rightarrow T}))))
\label{equ:f_loss}
\end{aligned}
\end{equation}
where $\sigma$ is the SoftMax function, $\mathds{1}$ is an indicator function and $\textbf{x}^{S_i\rightarrow T}_{i} = G_{ST}(E_i(\textbf{x}_{i}))$.The hyper-parameters $\lambda_4$ and $\lambda_5$ control the weights of the two different objective terms.

\subsection{MSGAN Learning}
\label{ssec:MSGANLearning}
We jointly solve the learning problems of the $MVAE_i$, $MVAE_T$, $GAN_S$, $GAN_T (i=1,2,\cdots,M)$, $F$ for the image reconstruction, the image translation, and the cycle-reconstruction pipelines and classification task.

Inheriting from GAN, training the proposed MSGAN framework results in solving a mini-max problem where the optimization aims to find a saddle point. 
It can be seen as a two player zero-sum game. 
The first player is a team consisting of the encoders and generators. 
The second player is a team consisting of the adversarial discriminators. 
In addition to defeating the second player, the first player has to minimize the MVAE losses and the cycle-consistency losses. 
We apply an alternating gradient update scheme similar to the one described in~\cite{goodfellow2014generative}.

The procedure is summarized in Algorithm~\ref{alg:Learning}, where $\bm{\alpha}_{S_1}$, $\cdots$, $\bm{\alpha}_{S_M}$, $\bm{\alpha}_{T}$,$\bm{\theta}_{ST}$, $\bm{\theta}_{TS}$, $\bm{\phi}_{S}$, $\bm{\phi}_{T}$, and $\bm{\varphi}$ are the parameters of $E_{S_1}$, $\cdots$, $E_{S_M}$, $G_{ST}$, $G_{TS}$, $D_S$, $D_T$, and $F$.

\begin{algorithm}[!t]
	{\small
		\KwIn{Sets of $M$ sources images with sentiment labels $(\textbf{x}_1, y_1)\in S_1$, $\cdots$, $(\textbf{x}_M, y_M)\in S_M$ , and target images $\textbf{x}_T\in \textbf{X}_T$, the maximum number of steps $T$}
		
		\KwOut{Predicted sentiment label of target image $\textbf{x}_T$}
		
		\For{$t\leftarrow 1\ \textbf{to}\ T$}{
			Sample a mini-batch of target images $\textbf{x}_T$;\\
			\For{$i\leftarrow 1\ \textbf{to}\ M$}{
				  Sample a mini-batch of source images $\textbf{x}_i$;\\
				   \tcc{Updating $\bm{\phi}_{S}$ , $\bm{\phi}_{T}$ when fixing $\bm{\alpha}_{S_i}$, $\bm{\alpha}_{T}$, $\bm{\theta}_{ST}$, $\bm{\theta}_{TS}$, and $\bm{\varphi}$}
				  
				  Compute $E_{S_i}(\textbf{x}_i;\bm{\alpha_{S_i}})$, $G_{ST}(\textbf{z}_i;\bm{\theta}_{ST})$ with current $\bm{\alpha_{S_i}}, \bm{\theta}_{ST}$;\\
				  Compute $E_{T}(\textbf{x}_T;\bm{\alpha_{T}})$, $G_{TS}(\textbf{z}_T;\bm{\theta}_{TS})$ with current $\bm{\alpha_{T}}, \bm{\theta}_{TS}$;\\
				  Update $\bm{\phi}_{S}$ and $\bm{\phi}_{T}$ by taking a Adam step on mini-batch loss  $\mathcal{L}_{GAN_i}$, $\mathcal{L}_{GAN_T}$ in
				  Eq.~(\ref{equ:gan_loss_s}) and Eq.~(\ref{equ:gan_loss_t}), respectively;

				  \tcc{Updating $\bm{\alpha}_{S_i}$, $\bm{\alpha}_{T}$, $\bm{\theta}_{ST}$ and $\bm{\theta}_{TS}$ when fixing  $\bm{\phi}_{T}$ , $\bm{\phi}_{S}$ and $\bm{\varphi}$}
				  Update $\bm{\alpha}_{S_i}$, $\bm{\alpha}_{T}$, $\bm{\theta}_{ST}$ and $\bm{\theta}_{TS}$ by taking a Adam step on mini-batch loss 
				  $\mathcal{L}_{MVAE-GAN}$, $\mathcal{L}_{ESC}$ in
				 Eq.~(\ref{equ:MVAE-GAN_loss}) and Eq.~(\ref{equ:esc_loss}), respectively;

				  \tcc{Updating $\bm{\alpha}_{S_i}$, $\bm{\alpha}_{T}$, $\bm{\varphi}$ when fixing $\bm{\phi}_{T}$ ,$\bm{\phi}_{S}$,$\bm{\theta}_{ST}$,and $\bm{\theta}_{TS}$}
				  Update $\bm{\alpha}_{S_i}$, $\bm{\alpha}_{T}$, $\bm{\varphi}$ by taking a Adam step on mini-batch loss $\mathcal{L}_{f}$ in Eq.~(\ref{equ:f_loss});

		}
		}
		\Return $E_T(\textbf{x}_T;\bm{\alpha_{T}})$, $F(\textbf{x}_T;\bm{\varphi})$.}
	
	\small\caption{Adversarial training procedure of the proposed MSGAN model}
	\label{alg:Learning}
\end{algorithm}

\begin{table*}[!t]
	\centering\small
	\caption{Classification accuracy (\%) comparison of the proposed MSGAN method to the state-of-the-art approaches. We separately use FI, Artphoto, Twitter \uppercase\expandafter{\romannumeral1} and Twitter \uppercase\expandafter{\romannumeral2} as the target domain and the rest three datasets as source domains. The best method trained on the source domains is emphasized in bold.}
	\begin{tabular}
		{c | c | c c c c }
		\hline
		Standards& Method & FI & Artphoto & Twitter \uppercase\expandafter{\romannumeral1} & Twitter \uppercase\expandafter{\romannumeral2} \\
		\hline
		\multirow{3}*{Source-only} & single-best  &  64.64 & 61.11  &  61.41 & 71.9  \\
		~&source-combined &   65.2  & 59.87  &  60.99 & 70.24\\
		~&resnet-simple-extend &  65.71   & 58.02  &  58.66 & 69.34 \\
		\hline
		\multirow{2}*{Single-best DA} & CycleGAN~\cite{zhu2017unpaired}  & 63.87 & 61.11  & 61.59  & 70.24\\
		~& CycleEmotionGAN~\cite{zhao2019cycleemotiongan}  & 67.48  &  61.72 &  62.38 & 71.9 \\
		\hline
		\multirow{2}*{Source-combined DA}   & CycleGAN~\cite{zhu2017unpaired}   &  66.05 &  60.49 &  59.84 & 71.07 \\
		~& CycleEmotionGAN~\cite{zhao2019cycleemotiongan}  &  67.78 &  61.96 &  61.77 & \textbf{72.72} \\
		\hline
		\multirow{2}*{Multi-source DA}   & DCTN~\cite{xu2018deep}   &  65.31 &  62.34 &  62.59 & 66.94 \\
		~& MDAN~\cite{zhao2018adversarial}  & 63.92  & 60.36  & 61.59 &71.9 \\
		~& MSGAN(Ours)  & \textbf{70.63} &  \textbf{63.58} &  \textbf{65.35} & \textbf{72.72} \\
		\hline
		\multicolumn{2}{c|}{Oracle (Train on sufficiently-labeled target data)}  &  75.24 &  64.81 &  68.5 & 72.72 \\
		\hline
	\end{tabular}
	\label{tab:exp}
\end{table*}

\section{Experiments}
\label{sec:Experiments}
This section presents the experimental analysis of MSGAN. First, the detailed experimental setups are introduced, including the datasets, baselines, and evaluation metrics. 
Second, the performance of MSGAN and the state-of-the-art algorithms in MDA is reported. 
Finally, an in-depth analysis on MSGAN, including a parameter sensitivity analysis and an ablation study, is presented.

\subsection{Datasets}
\label{ssec:Datasets}
We evaluate our framework on four public datasets including the Flickr and Instagram (\textbf{FI})~\cite{you2016building}, Artistic (\textbf{ArtPhoto}) dataset~\cite{machajdik2010affective}, \textbf{Twitter \uppercase\expandafter{\romannumeral1}}~\cite{you2015robust} and \textbf{Twitter \uppercase\expandafter{\romannumeral2}} ~\cite{you2016building} datasets.
FI is collected by querying with eight sentiment categories as keywords from social websites.
22,700 images are included in the FI dataset.
ArtPhoto dataset consists of 806 artistic photographs from a photo sharing site searched by emotion categories. 
We combine excitement, amusement, awe, contentment as positive images and disgust, anger, fear, sadness as negative images~\cite{mikels2005emotional}.
The Twitter \uppercase\expandafter{\romannumeral1} and Twitter \uppercase\expandafter{\romannumeral2} datasets are collected from the social websites and labeled with sentiment polarity (i.e. positive, negative) labels, which consist of 1,269 and 603 images, respectively.

\subsection{Baselines}
\label{ssec:Baselines}
To the best of our knowledge, MSGAN is the first work on multi-source domain adaptation for classifying visual sentiment. We compare MSGAN with three types of baseline algorithms, termed \textbf{Source-only}, \textbf{Single-source DA}, and \textbf{Multi-source DA}. The \textbf{Source-only} methods are trained on source images and directly test their classification performance on the target images. The \textbf{Single-source DA} methods include CycleGAN~\cite{zhu2017unpaired} and CycleEmotionGAN~\cite{zhao2019cycleemotiongan}.
For CycleGAN, we extend the original transfer network, 
\emph{i.e.} first adapt the source images to the adapted ones cycle-consistently, and then train the classifier on the adapted source images with the emotion labels from corresponding source images.
Since those methods perform in single-source setting, we employ two MDA standards: 
(1) single-best, \emph{i.e.} performing adaptation on each single source, and we choose single best performance from three source results; 
(2) source-combined, \emph{i.e.} all source domains are combined into a traditional single source. 
For resnet-simple-extend, our encoder and classifier can be seen as an simply extension of Resnet$-$18~\cite{he2016deep}. We train a classifier with the same network on the source combined. 
Additionally, we introduce two methods, DCTN~\cite{xu2018deep}, MDAN~\cite{zhao2018adversarial} as the \textbf{Multi-source DA} baselines.
For comparison, we also report the results of an oracle setting, where the classifier is both trained and tested both on the target domain.


\subsection{Comparison with State-of-the-art}
\label{ssec:comparison}
The performance comparisons between the proposed MSGAN model and the state-of-the-art approaches measured by classification accuracy are shown in Table~\ref{tab:exp}.
From the results, we have the following observations:

\textbf{(1)} The source-only methods can't handle the \emph{domain shift} or \emph{dataset bias}, where a rough combination is used to transfer all source data. 
Different domains represent the diverse joint probability distributions of observed images and emotion labels,so a simple combination of all source data for training is harmful to the model performance. 

\textbf{(2)} Both adaptation methods, CycleGAN~\cite{zhu2017unpaired} and CycleEmotionGAN~\cite{zhao2019cycleemotiongan}, are superior to the source-only methods, while CycleEmotionGAN performs better. 
This result demonstrates the effectiveness of CycleEmotionGAN for unsupervised domain adaptation in classifying image emotions. 
Obviously, in source-combined settings, adaption methods transfer the negative samples across the multiple source domains, which indicates multiple sources domain adapter should not be modeled in the same way with single source domain adapter.

\textbf{(3)} The proposed MSGAN achieves superior performance over the state-of-the-art approaches. 
The improvements benefit from four aspects: image reconstruction, image translation, cycle-reconstruction pipelines and the unified latent space. 
Firstly, compared with source-combined DA methods, our MSGAN further improves the classification performance, which demonstrates the proposed sentiment latent space can bridge the gap of multiple sources more effectively.
Especially, the images in Artphoto dataset, such as abstract oil painting, are far different from other source datasets. However, MSGAN can better distinguish them.
Thus, MSGAN further improves the classification performance than other adaption methods.
Secondly, compared to single-source DA, MSGAN utilizes more useful information from multiple sources.
Thirdly, while other multi-source DA methods only consider the alignment between each source domain and target domain, MSGAN attempts to align all source and target domains jointly.
In addition, existing DA methods, such as CycleGAN~\cite{zhu2017unpaired} and MDAN~\cite{zhao2018adversarial}, focus on matching the visual features but ignore the semantic labels. Therefore, they may not well characterize the consistency of visual sentiment. 
As a result, these methods may not preserve the mappings between visual content and the corresponding sentiment. 
However, all the DA methods with sentiment consistency loss, such as single-source method~\cite{zhao2019cycleemotiongan} and our multi-source method, significantly outperform the source-only approach, which demonstrates the effectiveness of preserving the sentiments of the adapted images for visual sentiment classification. 

\textbf{(4)} The oracle method, \emph{i.e.} testing on the target domain using the model trained on the same domain, achieves the best performance. 
However, this model is trained using the ground truth sentiment labels from the target domain, which are actually unavailable in unsupervised domain adaptation.

\begin{table}[!t]
	\centering\small
	\caption{Ablation study on different components in MSGAN. Baseline denotes using GAN with multiple encoders network, MVAE+GAN denotes using both MVAE loss and GAN loss, +CC denotes using the cycle-consistency loss, +ESC denotes using the emotional semantic consistency loss.}
	
	 \setlength{\tabcolsep}{1mm}{
	\begin{tabular}
		{c |  c c  }
		\hline
		Method & FI & Artphoto \\
		\hline
		baseline & 65.71 &  58.02 \\
		MVAE$+$GAN & 67.2  &  60.99  \\
		MVAE$+$GAN$+$CC & 69.03 & 62.16 \\
		MVAE$+$GAN$+$ESC  &  68.76 &  60.7  \\
		MVAE$+$GAN$+$ESC$+$CC & 70.63  & 63.58  \\
		\hline
	\end{tabular}}
	\label{tab:ablation}
\end{table}

\begin{figure*}[!t]
\begin{center}
\centering \includegraphics[width=0.91\linewidth]{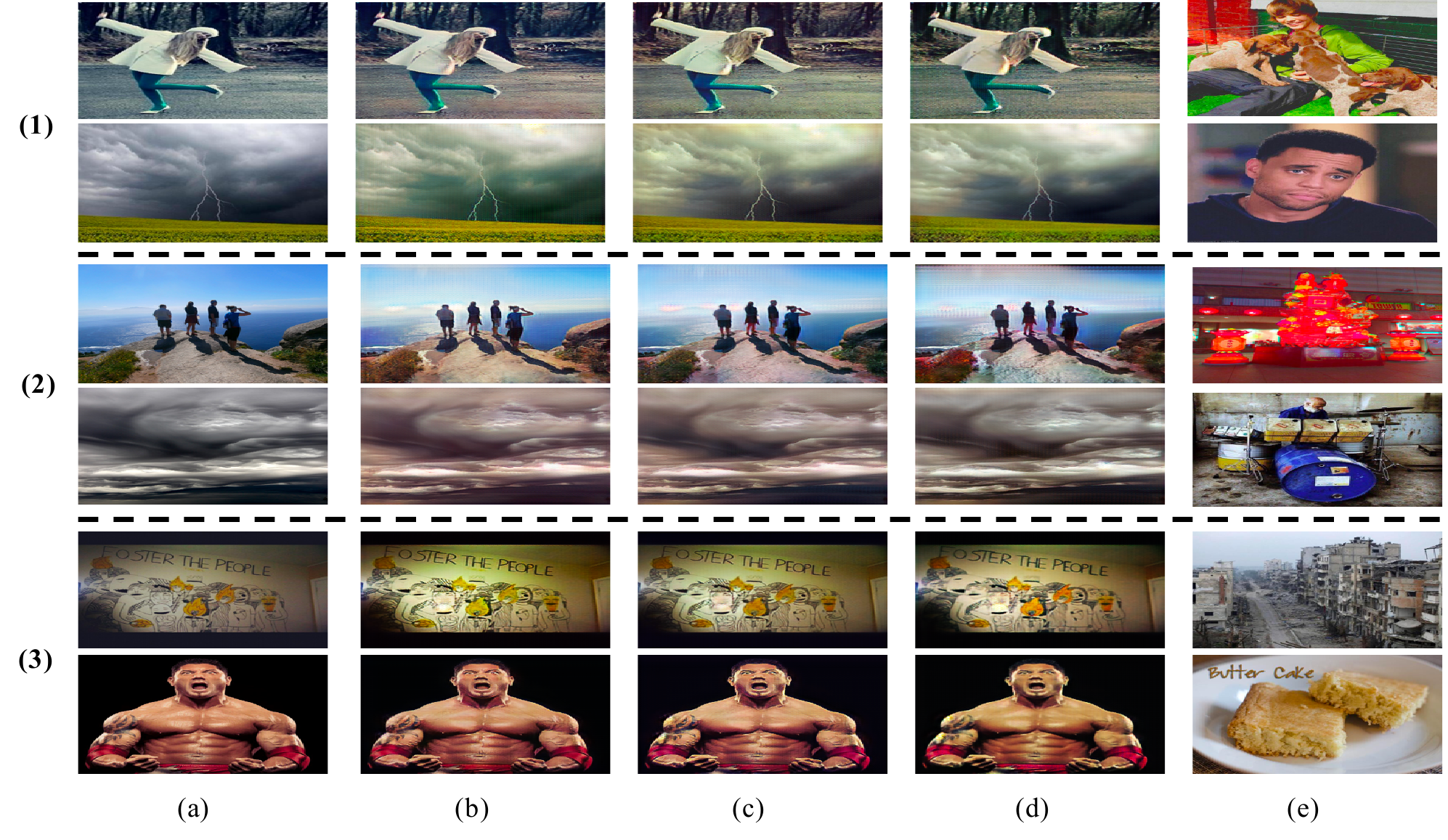}
\caption{
Visualization of image translation. From left to right are: (a) original source images, (b) MVAE+GAN, (c) MVAE+GAN+CC, (d) MVAE+GAN+CC+ESC, (e) target Twitter\uppercase\expandafter{\romannumeral1} image.
The source images of (1)(2)(3) are from Artphoto, FI, and Twitter\uppercase\expandafter{\romannumeral2} datasets respectively.
With adding the components of MSGAN, the adapted images look more visually similar to the target images than the source images.}
\label{fig:Vis}
\end{center}
\end{figure*}

\begin{figure}[!t]
	\begin{center}
		\subfigure[FI dataset as target]{
			\includegraphics[width=0.465\linewidth]{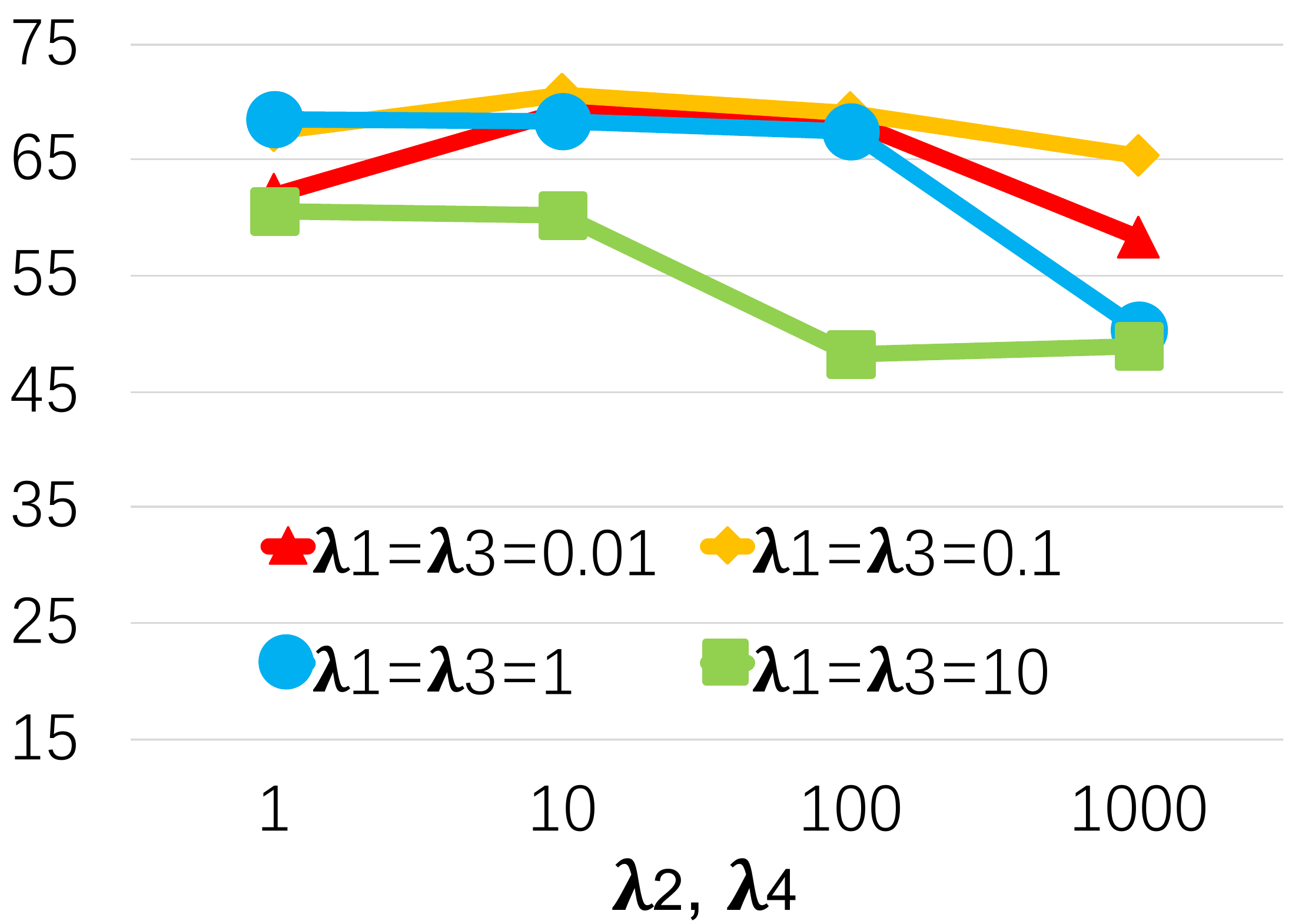}
		}
		\subfigure[Artphoto as target]{
			\includegraphics[width=0.465\linewidth]{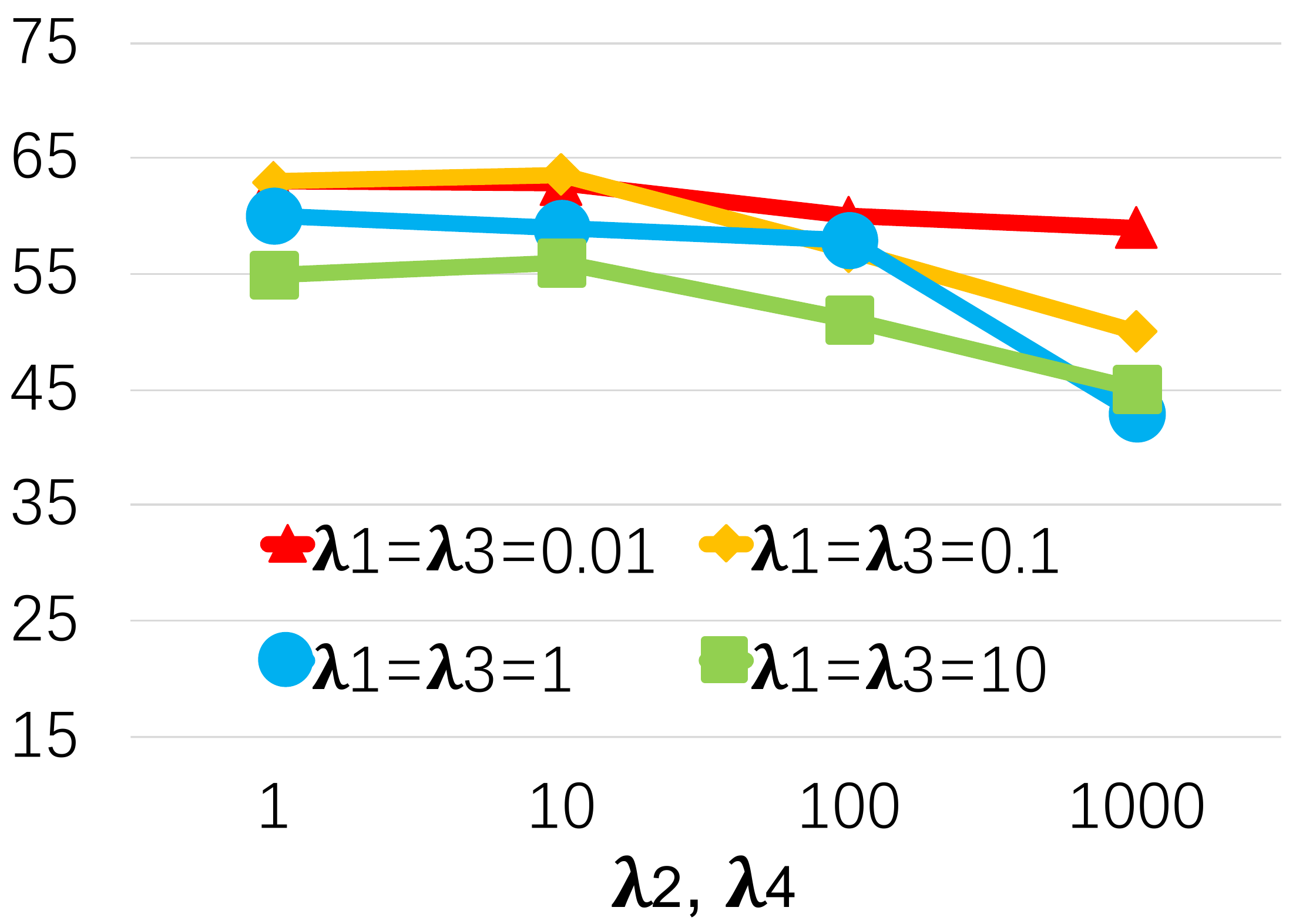}
		}
		\caption{Sensitivity of $\lambda$ in Eq.~(\ref{equ:mvae_loss_s})(\ref{equ:mvae_loss_t})(\ref{equ:cyc_loss_s})(\ref{equ:cyc_loss_t}) of the proposed MSGAN model.}
		\label{fig:ablation_1}
	\end{center}
\end{figure}

\noindent\textbf{Visualization}
We visualize the results of image-space adaptation from Artphoto, FI, Twitter \uppercase\expandafter{\romannumeral2} to Twitter \uppercase\expandafter{\romannumeral1} in Figure~\ref{fig:Vis}. 
We can see that with our final proposed MSGAN method (d), the styles of the images are similar to FI while the emotion semantic information is well preserved.

\subsection{Ablation Study}

We incrementally investigate the effectiveness of different components in MSGAN. The results are shown in Table~\ref{tab:ablation}. 
We can observe that: 
(1) MVAE+GAN can obtain better performance by making different adapted domains more closely aggregated; 
(2) adding the cycle-consistency loss could further improve the accuracy, again demonstrating the effectiveness of the unified sentiment latent space; 
(3) ESC loss also contributes to the visual sentiment adaptation task; 
(4) the modules are orthogonal to each other, since adding each one of them does not introduce performance degradation.

\textbf{Parameter Sensitivity.} we analyze the impact of the hyper-parameter values in Eq.~(\ref{equ:mvae_loss_s})(\ref{equ:mvae_loss_t})(\ref{equ:cyc_loss_s})(\ref{equ:cyc_loss_t}) to the sentiment classification accuracy. 
For different weight values on the negative log likelihood terms $(\emph{i.e.},\lambda_1, \lambda_3)$, we computed the achieved classification accuracy over different weight values on the KL terms $(\emph{i.e.},\lambda_0, \lambda_2)$ for both $\{$Artphoto, Twitter \uppercase\expandafter{\romannumeral1}, Twitter \uppercase\expandafter{\romannumeral2}$\}$ $\rightarrow$ FI and $\{$FI, Twitter \uppercase\expandafter{\romannumeral1}, Twitter \uppercase\expandafter{\romannumeral2}$\}$ $\rightarrow$ Artphoto. 
The results are reported in Figure~\ref{fig:ablation_1}. 
We can observe that, in general, a larger weight value on the negative log likelihood terms yields a better result. 
We also find that setting the weights of the KL terms to 0.1 can result in consistently good performance. We hence set $\lambda_0 = \lambda_2 = 10$, $\lambda_1 = \lambda_3 = 0.1$.

\section{Conclusion}
\label{sec:Conclusion}
In this paper, we tackle the problem of multi-source domain adaptation (MDA) in visual sentiment classification.
A novel framework, termed Multi-source Sentiment Generative Adversarial Network (MSGAN), is proposed to learn a unified sentiment latent space such that data from both the source and target domains share a similar distribution. 
Such mappings are learned via three pipelines, including image reconstruction, image translation, and cycle reconstruction. 
Extensive experiments conducted on four benchmark datasets demonstrate that MSGAN significantly outperforms the state-of-the-art MDA approaches for visual sentiment classification.
For further studies, we plan to extend the MSGAN model to other image emotion recognition tasks, such as emotion distribution learning~\cite{zhao2017continuous}. 
We will also investigate methods that can improve the intrinsic problem of GAN variants in training stability.

\section{Acknowledgement}
\label{sec:Acknowledgement}
NExT$++$ research is supported by the National Research Foundation, Prime Minister's Office, Singapore under its IRC@SG Funding Initiative.

\small\bibliographystyle{aaai}
\bibliography{MSGAN_Refe}

\begin{thebibliography}{}

\bibitem[\protect\citeauthoryear{Alaei, Becken, and
  Stantic}{2019}]{alaei2019sentiment}
Alaei, A.~R.; Becken, S.; and Stantic, B.
\newblock 2019.
\newblock Sentiment analysis in tourism: capitalizing on big data.
\newblock {\em JTR} 58(2):175--191.

\bibitem[\protect\citeauthoryear{Blitzer \bgroup et al\mbox.\egroup
  }{2008}]{blitzer2008learning}
Blitzer, J.; Crammer, K.; Kulesza, A.; Pereira, F.; and Wortman, J.
\newblock 2008.
\newblock Learning bounds for domain adaptation.
\newblock In {\em NIPS},  129--136.

\bibitem[\protect\citeauthoryear{Borth \bgroup et al\mbox.\egroup
  }{2013}]{borth2013large}
Borth, D.; Ji, R.; Chen, T.; Breuel, T.; and Chang, S.-F.
\newblock 2013.
\newblock Large-scale visual sentiment ontology and detectors using adjective
  noun pairs.
\newblock In {\em ACM MM},  223--232.

\bibitem[\protect\citeauthoryear{Ghifary \bgroup et al\mbox.\egroup
  }{2015}]{ghifary2015domain}
Ghifary, M.; Bastiaan~Kleijn, W.; Zhang, M.; and Balduzzi, D.
\newblock 2015.
\newblock Domain generalization for object recognition with multi-task
  autoencoders.
\newblock In {\em ICCV},  2551--2559.

\bibitem[\protect\citeauthoryear{Ghifary \bgroup et al\mbox.\egroup
  }{2016}]{ghifary2016deep}
Ghifary, M.; Kleijn, W.~B.; Zhang, M.; Balduzzi, D.; and Li, W.
\newblock 2016.
\newblock Deep reconstruction-classification networks for unsupervised domain
  adaptation.
\newblock In {\em ECCV},  597--613.

\bibitem[\protect\citeauthoryear{Goodfellow \bgroup et al\mbox.\egroup
  }{2014}]{goodfellow2014generative}
Goodfellow, I.; Pouget-Abadie, J.; Mirza, M.; Xu, B.; Warde-Farley, D.; Ozair,
  S.; Courville, A.; and Bengio, Y.
\newblock 2014.
\newblock Generative adversarial nets.
\newblock In {\em NIPS},  2672--2680.

\bibitem[\protect\citeauthoryear{He \bgroup et al\mbox.\egroup
  }{2016}]{he2016deep}
He, K.; Zhang, X.; Ren, S.; and Sun, J.
\newblock 2016.
\newblock Deep residual learning for image recognition.
\newblock In {\em CVPR},  770--778.

\bibitem[\protect\citeauthoryear{Katsurai and Satoh}{2016}]{katsurai2016image}
Katsurai, M., and Satoh, S.
\newblock 2016.
\newblock Image sentiment analysis using latent correlations among visual,
  textual, and sentiment views.
\newblock In {\em ICASSP},  2837--2841.

\bibitem[\protect\citeauthoryear{Liu and Tuzel}{2016}]{liu2016coupled}
Liu, M.-Y., and Tuzel, O.
\newblock 2016.
\newblock Coupled generative adversarial networks.
\newblock In {\em NIPS},  469--477.

\bibitem[\protect\citeauthoryear{Liu \bgroup et al\mbox.\egroup
  }{2017}]{liu2017unsupervised}
Liu, M.-Y.; Breuel, T.; Kautz; and Jan.
\newblock 2017.
\newblock Unsupervised image-to-image translation networks.
\newblock In {\em NIPS},  700--708.

\bibitem[\protect\citeauthoryear{Machajdik}{2010}]{machajdik2010affective}
Machajdik, Jana, H.~A.
\newblock 2010.
\newblock Affective image classification using features inspired by psychology
  and art theory.
\newblock In {\em ACM MM},  83--92.

\bibitem[\protect\citeauthoryear{Mansour, Mohri, and
  Rostamizadeh}{2009}]{mansour2009domain}
Mansour, Y.; Mohri, M.; and Rostamizadeh, A.
\newblock 2009.
\newblock Domain adaptation with multiple sources.
\newblock In {\em NIPS},  1041--1048.

\bibitem[\protect\citeauthoryear{Mikels \bgroup et al\mbox.\egroup
  }{2005}]{mikels2005emotional}
Mikels, J.~A.; Fredrickson, B.~L.; Larkin, G.~R.; Lindberg, C.~M.; Maglio,
  S.~J.; and Reuter-Lorenz, P.~A.
\newblock 2005.
\newblock Emotional category data on images from the international affective
  picture system.
\newblock {\em BRM} 37(4):626--630.

\bibitem[\protect\citeauthoryear{Peng \bgroup et al\mbox.\egroup
  }{2015}]{peng2015mixed}
Peng, K.-C.; Sadovnik, A.; Gallagher, A.; and Chen, T.
\newblock 2015.
\newblock A mixed bag of emotions: Model, predict, and transfer emotion
  distributions.
\newblock In {\em CVPR},  860--868.

\bibitem[\protect\citeauthoryear{Peng \bgroup et al\mbox.\egroup
  }{2018}]{peng2018moment}
Peng, X.; Bai, Q.; Xia, X.; Huang, Z.; Saenko, K.; and Wang, B.
\newblock 2018.
\newblock Moment matching for multi-source domain adaptation.
\newblock {\em arXiv:1812.01754}.

\bibitem[\protect\citeauthoryear{Riemer \bgroup et al\mbox.\egroup
  }{2019}]{riemer2018learning}
Riemer, M.; Cases, I.; Ajemian, R.; Liu, M.; Rish, I.; Tu, Y.; and Tesauro, G.
\newblock 2019.
\newblock Learning to learn without forgetting by maximizing transfer and
  minimizing interference.
\newblock {\em ICLR}.

\bibitem[\protect\citeauthoryear{Sun and Shi}{2013}]{sun2013bayesian}
Sun, S.-L., and Shi, H.-L.
\newblock 2013.
\newblock Bayesian multi-source domain adaptation.
\newblock In {\em ICMLC}, volume~1,  24--28.

\bibitem[\protect\citeauthoryear{Sun, Feng, and
  Saenko}{2017}]{sun2017correlation}
Sun, B.; Feng, J.; and Saenko, K.
\newblock 2017.
\newblock Correlation alignment for unsupervised domain adaptation.
\newblock In {\em Domain Adaptation in Computer Vision Applications}.
\newblock  153--171.

\bibitem[\protect\citeauthoryear{Torralba and
  Efros}{2011}]{torralba2011unbiased}
Torralba, A., and Efros, A.~A.
\newblock 2011.
\newblock Unbiased look at dataset bias.
\newblock In {\em CVPR},  1521--1528.

\bibitem[\protect\citeauthoryear{Xu \bgroup et al\mbox.\egroup
  }{2018}]{xu2018deep}
Xu, R.; Chen, Z.; Zuo, W.; Yan, J.; and Lin, L.
\newblock 2018.
\newblock Deep cocktail network: Multi-source unsupervised domain adaptation
  with category shift.
\newblock In {\em CVPR},  3964--3973.

\bibitem[\protect\citeauthoryear{Yang \bgroup et al\mbox.\egroup
  }{2017}]{yang2017joint}
Yang, J.; She, D.; Sun; and Ming.
\newblock 2017.
\newblock Joint image emotion classification and distribution learning via deep
  convolutional neural network.
\newblock In {\em IJCAI},  3266--3272.

\bibitem[\protect\citeauthoryear{Yang \bgroup et al\mbox.\egroup
  }{2018a}]{yang2018weakly}
Yang, J.; She, D.; Lai, Y.-K.; Rosin, P.~L.; and Yang, M.-H.
\newblock 2018a.
\newblock Weakly supervised coupled networks for visual sentiment analysis.
\newblock In {\em CVPR},  7584--7592.

\bibitem[\protect\citeauthoryear{Yang \bgroup et al\mbox.\egroup
  }{2018b}]{yang2018retrieving}
Yang, J.; She, D.; Lai, Y.; and Yang, M.-H.
\newblock 2018b.
\newblock Retrieving and classifying affective images via deep metric learning.
\newblock In {\em AAAI}.

\bibitem[\protect\citeauthoryear{Yang \bgroup et al\mbox.\egroup
  }{2018c}]{yang2018visual}
Yang, J.; She, D.; Sun, M.; Cheng, M.-M.; Rosin, P.~L.; and Wang, L.
\newblock 2018c.
\newblock Visual sentiment prediction based on automatic discovery of affective
  regions.
\newblock {\em IEEE TMM} 20(9):2513--2525.

\bibitem[\protect\citeauthoryear{You \bgroup et al\mbox.\egroup
  }{2015}]{you2015robust}
You, Q.; Luo, J.; Jin, H.; and Yang, J.
\newblock 2015.
\newblock Robust image sentiment analysis using progressively trained and
  domain transferred deep networks.
\newblock In {\em AAAI},  381--388.

\bibitem[\protect\citeauthoryear{You \bgroup et al\mbox.\egroup
  }{2016}]{you2016building}
You, Q.; Luo, J.; Jin, H.; and Yang, J.
\newblock 2016.
\newblock Building a large scale dataset for image emotion recognition: The
  fine print and the benchmark.
\newblock In {\em AAAI},  308--314.

\bibitem[\protect\citeauthoryear{You, Jin, and Luo}{2017}]{you2017visual}
You, Q.; Jin, H.; and Luo, J.
\newblock 2017.
\newblock Visual sentiment analysis by attending on local image regions.
\newblock In {\em AAAI},  231--237.

\bibitem[\protect\citeauthoryear{Zhao \bgroup et al\mbox.\egroup
  }{2017}]{zhao2017continuous}
Zhao, S.; Yao, H.; Gao, Y.; Ji, R.; and Ding, G.
\newblock 2017.
\newblock Continuous probability distribution prediction of image emotions via
  multi-task shared sparse regression.
\newblock {\em IEEE TMM} 19(3):632--645.

\bibitem[\protect\citeauthoryear{Zhao \bgroup et al\mbox.\egroup
  }{2018a}]{zhao2018adversarial}
Zhao, H.; Zhang, S.; Wu, G.; Moura, J.~M.; Costeira, J.~P.; and Gordon, G.~J.
\newblock 2018a.
\newblock Adversarial multiple source domain adaptation.
\newblock In {\em NIPS},  8559--8570.

\bibitem[\protect\citeauthoryear{Zhao \bgroup et al\mbox.\egroup
  }{2018b}]{zhao2018predicting}
Zhao, S.; Yao, H.; Gao, Y.; Ding, G.; and Chua, T.-S.
\newblock 2018b.
\newblock Predicting personalized image emotion perceptions in social networks.
\newblock {\em IEEE TAFFC} 9(4):526--540.

\bibitem[\protect\citeauthoryear{Zhao \bgroup et al\mbox.\egroup
  }{2018c}]{zhao2018emotiongan}
Zhao, S.; Zhao, X.; Ding, G.; and Keutzer, K.
\newblock 2018c.
\newblock Emotiongan: unsupervised domain adaptation for learning discrete
  probability distributions of image emotions.
\newblock In {\em ACM MM},  1319--1327.

\bibitem[\protect\citeauthoryear{Zhao \bgroup et al\mbox.\egroup
  }{2019a}]{zhao2019multi}
Zhao, S.; Li, B.; Yue, X.; Gu, Y.; Xu, P.; Hu, R.; Chai, H.; and Keutzer, K.
\newblock 2019a.
\newblock Multi-source domain adaptation for semantic segmentation.
\newblock In {\em NIPS},  7285--7298.

\bibitem[\protect\citeauthoryear{Zhao \bgroup et al\mbox.\egroup
  }{2019b}]{zhao2019cycleemotiongan}
Zhao, S.; Lin, C.; Xu, P.; Zhao, S.; Guo, Y.; Krishna, R.; Ding, G.; and
  Keutzer, K.
\newblock 2019b.
\newblock Cycleemotiongan: Emotional semantic consistency preserved cyclegan
  for adapting image emotions.
\newblock In {\em AAAI},  2620--2627.

\bibitem[\protect\citeauthoryear{Zhu \bgroup et al\mbox.\egroup
  }{2017}]{zhu2017unpaired}
Zhu, J.-Y.; Park, T.; Isola, P.; and Efros, A.~A.
\newblock 2017.
\newblock Unpaired image-to-image translation using cycle-consistent
  adversarial networks.
\newblock In {\em ICCV},  2242--2251.

\bibitem[\protect\citeauthoryear{Zhuo \bgroup et al\mbox.\egroup
  }{2017}]{zhuo2017deep}
Zhuo, J.; Wang, S.; Zhang, W.; and Huang, Q.
\newblock 2017.
\newblock Deep unsupervised convolutional domain adaptation.
\newblock In {\em ACM MM},  261--269.

\end{thebibliography}

\end{document}